\documentclass{article}
\pdfoutput=1

\usepackage{arxiv}
\usepackage[export]{adjustbox}
\usepackage[para,online,flushleft]{threeparttable}
\usepackage{booktabs}
\usepackage{threeparttable}
\usepackage{makecell}
\usepackage[utf8]{inputenc} 
\usepackage[T1]{fontenc}    
\usepackage{hyperref}       
\usepackage{url}            
\usepackage{booktabs}       
\usepackage{amsfonts}       
\usepackage{nicefrac}       
\usepackage{microtype}      
\usepackage{graphicx}
\usepackage[caption=false]{subfig}
\usepackage{doi}
\usepackage{amsmath}
\usepackage{multirow}
\usepackage{multicol}
\usepackage{titlesec}
\titlespacing*{\section}{0pt}{0.3\baselineskip}{0.3\baselineskip}
\titlespacing*{\paragraph}{0pt}{0.3\baselineskip}{0.3\baselineskip}

\title{Uncertainty Quantification in Table Structure Recognition}

\author{Kehinde Ajayi\\
	Old Dominion University\\
	Norfolk, VA 23529 \\
	\texttt{kajay001@odu.edu} \\
    \And
    Leizhen Zhang\\
    Old Dominion University\\
    Norfolk, VA 23529 \\
    \texttt{lzhan011@odu.edu} \\
   \And
    Yi He\\
    Old Dominion University\\
    Norfolk, VA 23529\\
    \texttt{yihe@cs.odu.edu} \\
	\And
    Jian Wu\\
	Old Dominion University\\
	Norfolk, VA 23529 \\
	\texttt{j1wu@cs.odu.edu} \\
}

\renewcommand{\shorttitle}{Uncertainty Quantification in Table Structure Recognition}

\begin{document}
\maketitle

\begin{abstract}
Quantifying uncertainties for machine learning models is a critical step to reduce human verification effort by detecting predictions with low confidence. This paper proposes a method for uncertainty quantification (UQ) of table structure recognition (TSR). The proposed UQ method is built upon a mixture-of-expert approach termed Test-Time Augmentation (TTA). Our key idea is to enrich and diversify the table representations, to spotlight the cells with high recognition uncertainties. To evaluate the effectiveness, we proposed two heuristics to differentiate highly uncertain cells from normal cells, namely, masking and cell complexity quantification. Masking involves varying the pixel intensity to deem the detection uncertainty. Cell complexity quantification gauges the uncertainty of each cell by its topological relation with neighboring cells. The evaluation results based on standard benchmark datasets demonstrate that the proposed method is effective in quantifying uncertainty in TSR models. To our best knowledge, this study is the first of its kind to enable UQ in TSR tasks. Our code and data are available at: https://github.com/lamps-lab/UQTTA.git.
\end{abstract}


\section{Introduction}
Table recognition has been studied in recent years to facilitate document understanding and retrieval tasks \cite{Hashmi_2021}. This task can be decomposed into two subtasks: table detection (TD) and table structure recognition (TSR). TD aims to automatically identify tables present in digital documents. TSR aims to identify the rows, columns, and individual text cells in table images. Early works use classical machine learning models such as conditional random fields \cite{wei2006table} and hidden markov models \cite{babatunde2015automatic}. Recently, deep learning approaches e.g., \cite{prasad2020cascadetabnet,Graph-based-TSR-lee,splerge} have been proposed. The output files for TSR models typically contain the coordinates of the identified cells in terms of row and column numbers and coordinates of bounding boxes that enclose the cells, so the cell content can be recognized by subsequent optical character recognition (OCR) software. 

The current TSR models can automatically identify cell locations, but the results do not predict uncertainties \cite{ReS2TIM,splerge}. This prevents TSR models to be applied in real-world scenarios, such as faithfully extracting domain tabular data for downstream analysis in scientific domains, e.g., materials science. It is costly and sometimes infeasible for domain experts to verify all data extracted by machine learning models. Therefore, automatically quantifying TSR uncertainties is crucial to minimize human effort for data verification.

UQ methods have been proposed for deep learning-based solutions of several natural language processing \cite{Xiao_2019} and computer vision tasks \cite{Shen_2020}, but to our best knowledge, it has not been incorporated into TSR. A recent work \cite{prasad2020cascadetabnet} attempted to incorporate confidence estimation into the cell structures of the tables detected in document images but the confidence scores were represented as binaries indicating whether a cell was detected or not. Our work will abridge the gap by quantifying uncertainties for TSR models as continuous values.

Uncertainties in a machine learning model can arise from two major sources, namely, aleatoric uncertainty (also known as data uncertainty) and epistemic uncertainty (also known as model uncertainty) \cite{abdar2021review}. Aleatoric uncertainty occurs as a result of measurement noise, data missingness, or outliers \cite{Chang_2020}, while epistemic uncertainty occurs due to the choice of model architecture, hyperparameters, and initialization \cite{gal2016dropout}. Several methods have been proposed to quantify the uncertainties in deep learning-based models, such as Bayesian methods \cite{mullachery2018bayesian}, Monte Carlo (MC) dropout \cite{gal2016dropout}, and Ensembles \cite{NEURIPS2021_a70dc404}. Bayesian neural networks use prior distributions to represent prior beliefs about the parameters of the neural networks, which are updated based on the data during training \cite{mullachery2018bayesian}. Once the model is trained, posterior distributions can be used to estimate uncertainty. MC dropout is based on dropout regularization \cite{gal2016dropout}. Specifically, during prediction, the dropout is applied multiple times to the network, and the variance of the predictions can be used as a measure of uncertainty. An ensemble method involves training multiple models with different architectures and combining their predictions \cite{NEURIPS2021_a70dc404}. The variance of the ensemble predictions can be used as a measure of uncertainty. Although Bayesian or MC dropout methods are easier to interpret, they are hard to scale up because they require multiple forward passes through the network for each prediction and require modifications to the neural network architecture to incorporate uncertainty \cite{gal2016dropout}. Ensemble methods are more scalable, robust, and flexible because they are agnostic to neural network architecture \cite{NEURIPS2021_a70dc404}. 

Our proposed UQ pipeline adopts the Test-Time Augmentation (TTA), a technique that involves applying data augmentation to samples during inference (or testing) time and then ensembling the predictions \cite{wang2018test}.

One key component of the vanilla TTA is data augmentation, which is usually task-dependent. The augmentation methods should be controllable and orthogonal. For TSR, we explore four heuristic methods to augment test table images, which were chosen to probe the differential performance of the pre-trained model using variations of the test data. We combined the original and augmented images to obtain ensemble results, at different confidence levels at a certain Intersection over Union (IoU) threshold, computed by dividing the area of intersection between the predicted bounding boxes (bboxes) and the ground truth bboxes by the area of the union of the two bboxes. If the predicted IoU is greater than the threshold, the two bboxes are deemed to match.

One challenge in evaluating UQ methods is the lack of human-labeled ground truth. Therefore, we proposed two heuristics (1) masking, which involves varying the pixel intensities of table images and then quantifying uncertainties using confidence level estimation, and (2) cell complexity quantification, which models the complexity of cell relations using undirected subgraphs and uses the complexity of subgraphs as a surrogate for recognition uncertainty.

To showcase the efficacy of our UQ model, we apply it to CascadeTabNet \cite{prasad2020cascadetabnet}, a recently proposed TSR model, which was retrainable.  However, our UQ framework can be integrated into other retrainable TSR models. The contributions of this paper are below:
\begin{enumerate}
  \item We proposed a novel ensemble method called \texttt{TTA-m} to quantify uncertainties for the results of TSR tasks and showcased its efficacy on a reproducible TSR framework called CascadeTabNet. 
  \item We proposed two controllable and scalable methods, masking and cell complexity quantification, to build the ground truth uncertainties. 
  \item We created a new dataset containing table images based on the ICDAR-19 document TSR competition. The new dataset augmented the original data using four heuristics and the ground truth uncertainties based on the two methods above. 
\end{enumerate}

\section{Related Work}

\subsection{Uncertainty Quantification}
UQ has been an area of interest in both traditional machine learning and deep learning \cite{abdar2021review}. Several methods have been proposed to quantify uncertainties in deep learning models, such as Bayesian models, Monte-Carlo Dropout, and Ensembles \cite{gal2016dropout}. 
One ensemble method is Bayesian model averaging \cite{NEURIPS2021_a70dc404}, which quantifies uncertainties by averaging predictions from multiple deep learning models training with augmented data.
TTA emerged as a straightforward method to enhance ensemble models \cite{ashukha2020pitfalls}. TTA is easier to implement and more computationally efficient. 

\subsection{Table Structure Recognition}
TSR has experienced significant strides recently due to the utilization of deep neural networks.
For example, Schreiber et al. \cite{schreiber2017deepdesrt} devised an innovative strategy amalgamating Faster R-CNN and Fully-Convolutional Network architectures. This convergence facilitated proficient table detection and precise cell position localization.
Siddiqui et al. \cite{deeptabstr} treated table images as comprehensive scenes using deformable convolution operations. This holistic approach offers a fertile ground for imbuing UQ into the fabric of scene-based representation.
The work of Xue et al. \cite{ReS2TIM} ventured into the realm of graph-based inference to unravel table syntactic structures using a cell relationship network.
Khan et al. \cite{GRU-TSR} harnessed bi-directional Gated Recurrent Units to discern intricate row and column boundaries in tables.
The split and merge model proposed by \cite{splerge} introduces a strategy for addressing TSR through hierarchical decomposition.
Lee et al. \cite{Graph-based-TSR-lee} framed TSR as a challenge of table graph reconstruction.
Hashmi et al. \cite{guided} implemented Mask R-CNN for anchor estimation in TSR.

However, most existing models are not able to quantify uncertainties of their predictions.

\begin{figure*}
    \centering
    \includegraphics[width=0.95\textwidth]{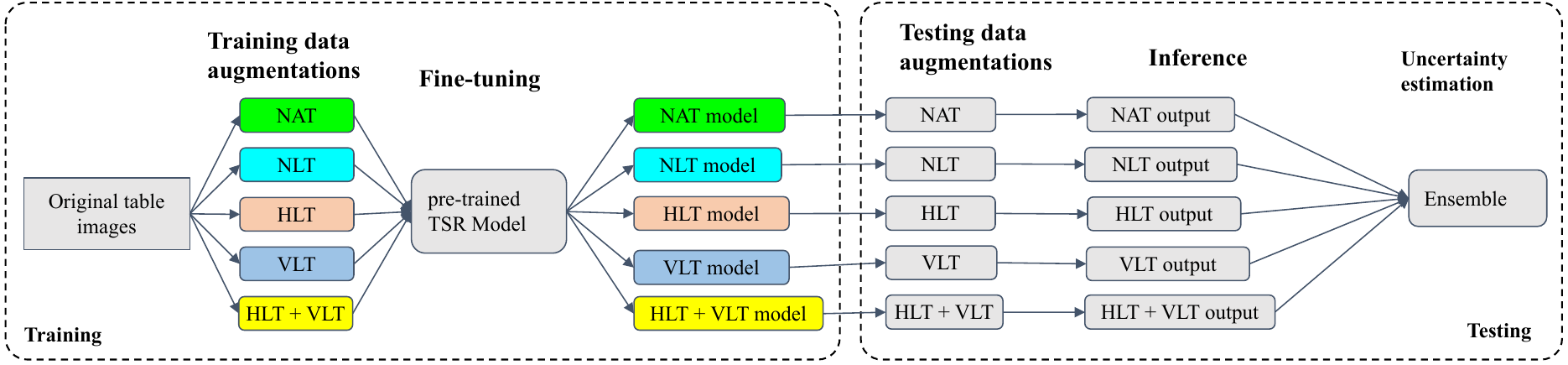}
     \caption{A schematic illustration of the proposed UQ pipeline (\texttt{TTA-m}). In the training phase, we fine-tuned the TSR model on the original tables and augmented tables. In the test phase, each model makes a prediction on table images similar to what it was trained on and then ensembling is applied on the model outputs. NAT: Non-Augmented Tables, NLT: No Lines Tables, HLT: Horizontal Lines Tables, VLT: Vertical Lines Tables.}
     \label{fig:uq_pipeline}
\end{figure*}

\section{TTA-m: Proposed UQ Pipeline}
Figure~\ref{fig:uq_pipeline} illustrates the architecture of the proposed UQ pipeline. The key modules include (1) Training with data augmentation, (2) Fine-tuning a pre-trained TSR model (using CascadeTabNet as a case study), (3) Inference with fine-tuned models, and (4) Uncertainty estimation. Because the proposed model modified the traditional TTA model, we call it \texttt{TTA-m}. For convenience, we define $M$ as the number of augmentation methods applied to the original data. 

\subsection{Data Augmentation}
\begin{figure}
    \centering
    \includegraphics[width=0.5\textwidth]{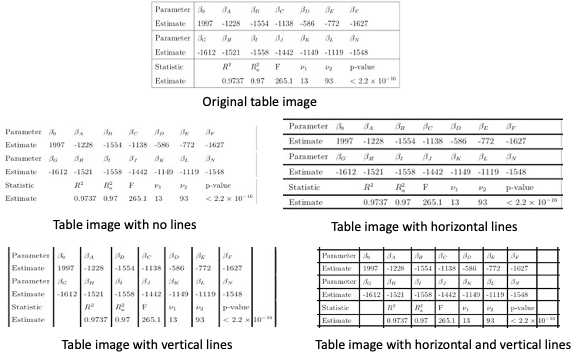}
     \caption{Augmentation examples of a table image.}
     \label{fig:aug}
\end{figure}

Data augmentation has become a practice for developing robust and transformation-resistant models \cite{mikolajczyk2018data}. We applied a combination of $M=4$ distinct data augmentation methods across the training and test stages. These methods encompass the removal of all lines (NLT), the addition of horizontal lines (HLT), the inclusion of vertical lines (VLT), and the incorporation of both horizontal and vertical lines (HLT + VLT). Figure~\ref{fig:aug} shows augmented table image examples. 

\subsection{Fine-tuning A Pre-trained TSR Model}
Instead of training the model on all augmented data, we fine-tuned a pre-trained TSR model on each set of augmented table images plus the original table images, resulting in $M+1$ distinct models.

\subsection{Predictions With Fine-tuned Model}
In the inference stage, we first applied the same augmentations to the \emph{test} set. Then, instead of evaluating the pre-trained model on the $M+1$ sets of table images, each fine-tuned model was applied to its corresponding test data set. For example, the model fine-tuned on tables with only vertical lines was applied on tables with vertical lines in the test set. Each fine-tuned model can be thus evaluated on standard binary classification metrics (precision, recall, and F1-score). These predictions are intermediate results. The final output is generated by the ensemble module.

\subsection{Uncertainty Estimation via Ensembles}\label{sec:ensembles}
In this module, we use an ensemble method to aggregate the predictions by fine-tuned models on augmented testing data. 
Our method is different from the traditional TTA because the training data is also augmented and the TSR model is fine-tuned before it is applied to the corresponding augmented test data. The uncertainty is modeled as the dispersion of the predicted results. 
The uncertainty estimation process involves progressively combining predictions from $M+1$ models based on the degree of overlap between cell predictions. This aggregation results in a set of merged cells with associated confidence scores, which collectively constitute the output.
The steps are detailed below. Here, we use $\theta_0$ to represent the IoU threshold as the criteria to match the bounding boxes of two predicted cells. 

\begin{figure}[h!]
    \centering
    \includegraphics[scale=0.54, center]{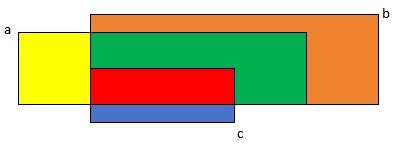}
     \caption{A schematic illustration of how to calculate confidence scores using bounding boxes predicted by three models (a, b, and c). Red color: 3/3 = 100\% confidence, Green color: 2/3 =  66.7\% confidence, Other colors: 1/3 = 33.3\% confidence.}
     \label{fig:conf_calc}
\end{figure}

\begin{enumerate}
    \item Obtain all the predicted bounding boxes from a randomly chosen model out of the $M+1$ models (considered as the base model).
    \item Obtain predicted bounding boxes from the second model.
    \item Calculate the IoU for each predicted cell from the base model against each from the second model. If the $\mbox{\rm IoU}\geq\theta_0$, merge these two cells and remove the second model's cell from its list.
    \item Repeat steps 2 and 3 for predictions from the remaining models $i=3, 4, ..., M+1$.

    \item Sequentially use $i = 2, 3, 4, \cdots, M+1$ models as new base models and perform calculations similar to Steps 1 to 4 for the cells that have not been merged.

    \item For all cell combinations generated in the above steps:
    \begin{enumerate}
        \item Count the number of distinct models contributing to each cell combination.
        \item Divide the count by $M+1$ to calculate the confidence score for that combination (Figure~\ref{fig:conf_calc}).
    \end{enumerate}
\end{enumerate}

\section {Evaluation}
Because the ground truth uncertainties are not available for the annotated cells, we proposed two methods to gauge uncertainties in different scenarios and use them as surrogates to evaluate the UQ pipeline.

\paragraph{Masking} The masking technique artificially changes the level of difficulty by varying the intensity of pixels on table images. Firstly, we increased the pixel intensity by a factor of 2 for each cell which makes the pixels appear fainter, and calculated the confidence estimates at the previously defined confidence scores. Next, we multiplied all pixel values by 3. If the pixel value becomes greater than 255, we set it to 255.
Then, we estimated the confidence scores of the TSR models at each intensity level. Our results indicate that the intensity of cell pixel values remarkably affects the distribution of confidence scores.

\begin{figure}[h!]
    \centering
    \includegraphics[width=0.47\textwidth]{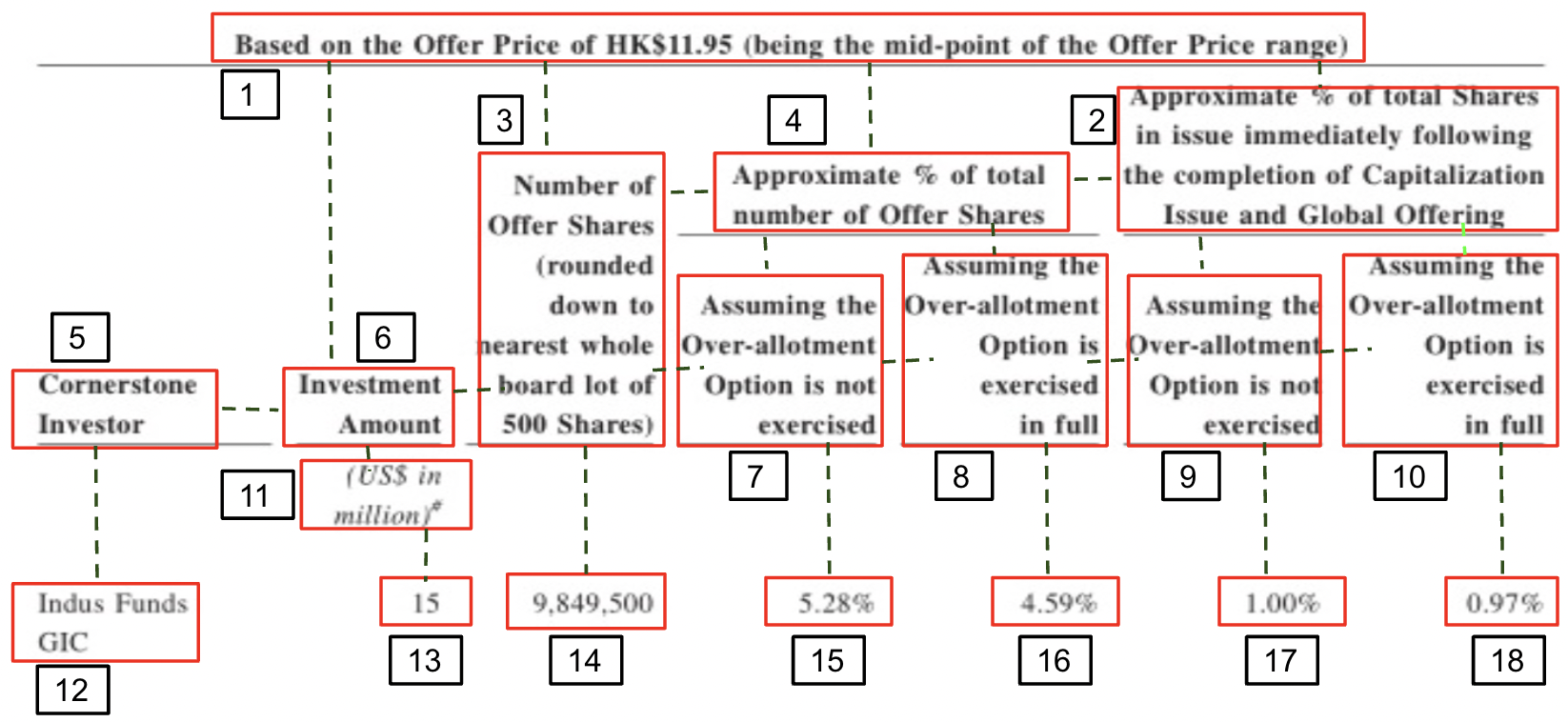}
     \caption{An example of the graph model of a table. Each cell is enclosed by a red box, with an ID labeled next to it. The dashed lines represent the connections of a cell to its adjacency cells and can be used for counting the adjacency degrees of a cell. For instance, cell 5 is connected by 2 green lines, so it has an adjancency degree of 2.}
     \label{fig:degree}
\end{figure}

\paragraph{Cell complexity quantification.}
We observed that TSR models are more likely to make mistakes for tables with complex structures. Specifically, table images may contain cells that span across multiple rows and/or columns, which are challenging for TSR models in general. To quantify the structure complexity, we model a table as a non-directed graph in which the nodes represent table cells, and the edges represent adjacency between cells. We considered four types of adjacency relations: left, top, right, and bottom. We define adjacency degree as the number of adjacency cells of the target cell, where a cell represents a unit within a table that contains meaningful text contents. Intuitively, the higher the degree of a cell is, the more likely the bounding box is incorrectly predicted. Therefore, our evaluation will test whether the fraction of cells detected with low confidence increases with the average degree of a table. Figure~\ref{fig:degree} illustrates examples of relationships that could exist between the cells of a table image. For evaluation, we manually annotated the relations between cells and constructed a graph for each table in the test set.

\section{Experimental Setup}
\subsection{Data}
We used the dataset created for the competition at the International Conference on Document Analysis and Recognition (ICDAR) 2019 containing real-world table images. For an even comparison, 
We used the dataset adopted by Prasad et al. \cite{prasad2020cascadetabnet} consisting of 543 table images selected from the benchmark dataset created for the competition at the International Conference on Document Analysis and Recognition (ICDAR-19). We randomly selected 443 table images for training and the remaining 100 table images for testing. The ICDAR 2019 dataset was originally used for both TD and TSR. We only used the ground truth labels for TSR.

\subsection{Baseline Methods}
We compare \texttt{TTA-m} with three baseline methods, including two variants of TTA and an active learning model.

\paragraph{TTA-t} TTA-t adds a small cell filter to the vanilla TTA to exclude small cells predicted by fine-tuned models. In observation, these small cells are usually produced by fine-tuned models using augmented data, and the large cells are produced by the model fine-tuned using the original data. These small cells occupy areas much smaller than the actual cell and thus do not contribute to confidence scores. Therefore, we crafted simple heuristics to remove them before ensembling the predicted results. 
The small cell filter is applied if its area ($S$) meets the following conditions. (1) the smaller cell is fully inside a bigger cell, $(S_{\rm small}\cap S_{\rm large})/S_{\rm small} = 1$;  (2) the smaller cell is significantly smaller than the bigger cell, or $S_{\rm small}/S_{\rm large}\leq0.5$.

\begin{figure}
    \centering
    \includegraphics[width=0.48\textwidth]{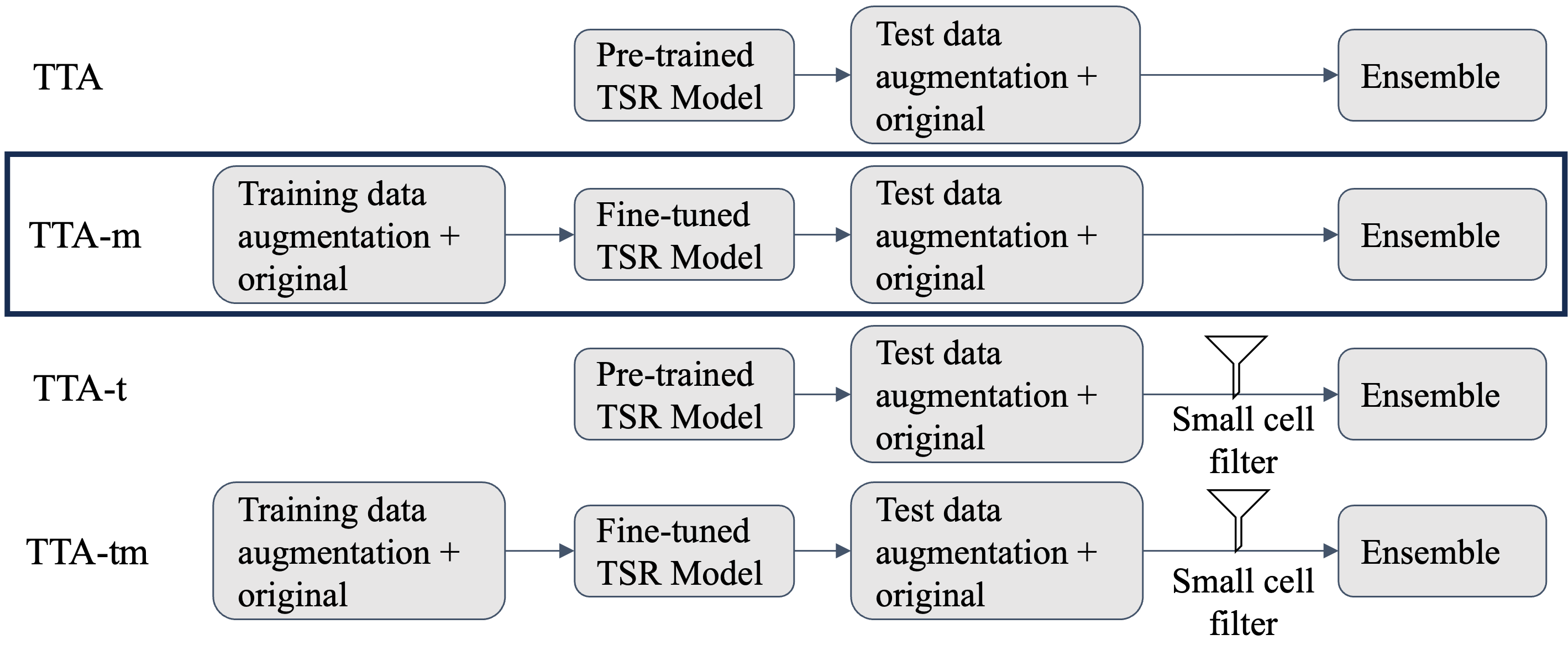}
    \caption{A schematic comparison of TTA variants implemented by this paper. TTA-m is proposed for its highest F1 over the others (Table~\ref{tab:retrain}).}
    \label{fig:allmodels}
\end{figure}

\paragraph{TTA-tm} {TTA-tm} combines the \texttt{TTA-t} and \texttt{TTA-m} model, which includes both training data augmentation and the small cell filter (Figure~\ref{fig:allmodels}).

\paragraph{Active Learning} We also compare our method against an active learning model proposed by Choi et al. \cite{choi2021active}. This method aims to reduce labeling costs by selecting only the most informative samples in a dataset. It uses a mixture density network that estimates a probabilistic distribution for each localization and classification head's output to explicitly estimate the aleatoric and epistemic uncertainty in a single forward pass of a single model. This method uses a scoring function that aggregates these uncertainties for both heads to obtain every image's informativeness score. We fine-tuned the baseline model on the table images in our training set and tested it on the original table images.

\subsection{Experiment Design}
We conduct three experiments to evaluate the proposed model. Our goal is to demonstrate that the ensemble results generated by the proposed model can accurately detect cells and that the confidence levels can be reliably used as a measure of uncertainty. To showcase the efficacy, we adopt CascadeTabNet \cite{prasad2020cascadetabnet} as the TSR model. All experiments are run on a server with 24 Intel Xeon cores, 384GB RAM, and 4x Nvidia 2080 Ti GPUs. The fine-tuning was run 1 time for either the training or the test set.

\paragraph{Experiment~1: Cell Detection.} In the first experiment, we compare model performances on cell recognition. The results in Table~\ref{tab:retrain} show that the \texttt{TTA-m} model outperforms all the baseline models in terms of the F1-scores. Note that we should not compare against F1-scores of models based on individual augmentation methods because they are not the final output of the pipeline. Notably, the employment of the ensemble technique resulted in a reduction in precision but an improvement in recall. However, the poor performance of the fine-tuned model on the augmented table images implies its low generalization capability to such augmented testing data. 

\begin{table}
 \caption{Comparing the models used in our study. The models compared include the original CascadeTabNet, baseline, and the fine-tuned CascadeTabNet on the four augmentation types and original table images.}
    \centering
    \begin{tabular}{c|c|c|c|c}
    \toprule
      {\bf Model} & {\bf Augmentation Method}  & {\bf Precision} & {\bf Recall} & {\bf F1} \\
    \midrule
   {} & Vertical lines  & 0.765 & 0.713 & 0.738 \\
   {} & Horizontal lines & 0.792 & 0.707 & 0.749 \\
    \textbf{TTA} & Both lines & 0.725 & 0.677 & 0.701 \\
    {} & No lines & 0.846 & 0.742 & 0.793 \\
    {} & {Original} & {0.883} & {0.767} & 0.823 \\
    \midrule
   \multicolumn{2}{c|}{TTA ensemble result}  & 0.683 & {0.824} & 0.753 \\
   \midrule
   \midrule
   {} & Vertical lines  & 0.854 & 0.755 & 0.802  \\
    {} & Horizontal lines & 0.841 & 0.758 & 0.798 \\
    \textbf{TTA-m} & Both lines & 0.844 & 0.744 & 0.791  \\
    \textbf{} & No lines & 0.838 & 0.738 & 0.785  \\
   \textbf{} & {Original} & {0.883} & 0.771 & {0.823} \\
   \midrule
    \multicolumn{2}{c|}{TTA-m ensemble result} & 0.761 & {0.835} & \textbf{0.798} \\
    \midrule
    \multicolumn{2}{c|}{TTA-tm ensemble result} & 0.778 & 0.831 & 0.806 \\
   \midrule
   \midrule
    \textbf{Baseline} & {Original} & {0.899} & 0.659 & 0.76 \\
    \bottomrule
    \end{tabular}
    \label{tab:retrain}
\end{table}

\begin{figure}[h!]
    \centering
    \includegraphics[scale=0.25, center]{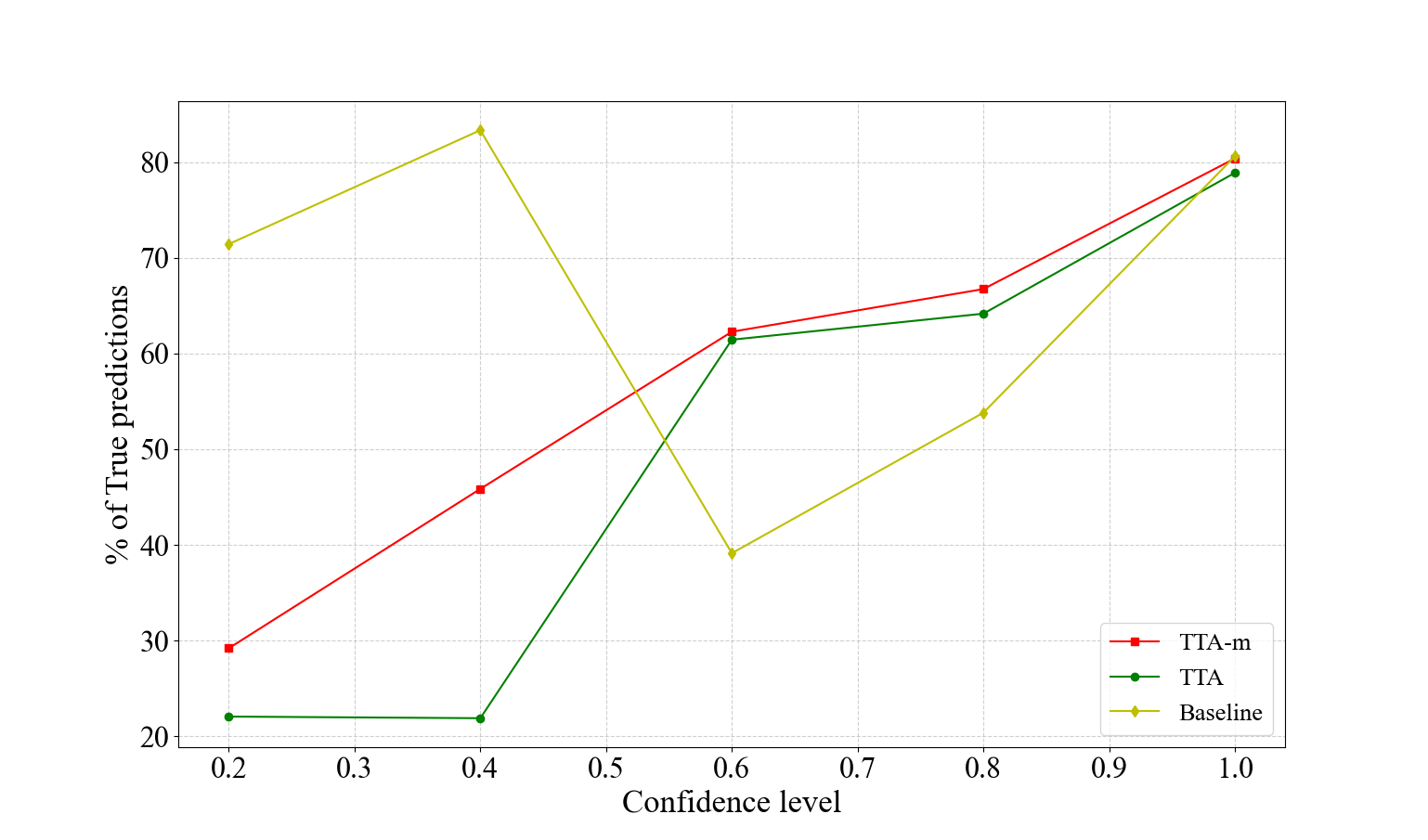}
     \caption{The percentage of true predictions for the TTA (green) and TTA-m (red) and baseline (light-green). TTA-m outperforms both the TTA and baseline models for confidence levels above 0.6.}
     \label{fig:tta-vs-ttam}
\end{figure}

\paragraph{Experiment 2: Confidence Level as a Measure of Uncertainty.} To demonstrate the confidence level output by the proposed model can be used as a measure of uncertainty, we calculated the percentage of correctly predicted cells at 0.2, 0.4, 0.6, 0.8, and 1.0 confidence levels and an IoU threshold $\theta_0=0.5$. The confidence levels were obtained based on the overlap area of the 5 predicted bounding boxes. For example, if only a cell is overlapped by two predicted bounding boxes, the confidence level is 2/5 = 0.4 (as shown in Section~\ref{sec:ensembles}).
Figure~\ref{fig:tta-vs-ttam} shows that the percentages of the correctly predicted cells increase monotonously with the confidence levels for the \texttt{TTA} and \texttt{TTA-m} models. However, this trend is not seen for the results produced by the baseline method. Figure~\ref{fig:tta-vs-ttam} also indicates that the \texttt{TTA-m} method correctly predicted over 80\% of cells with a confidence of 1.0.

\begin{figure}
    \centering
    \includegraphics[width=0.47\textwidth]{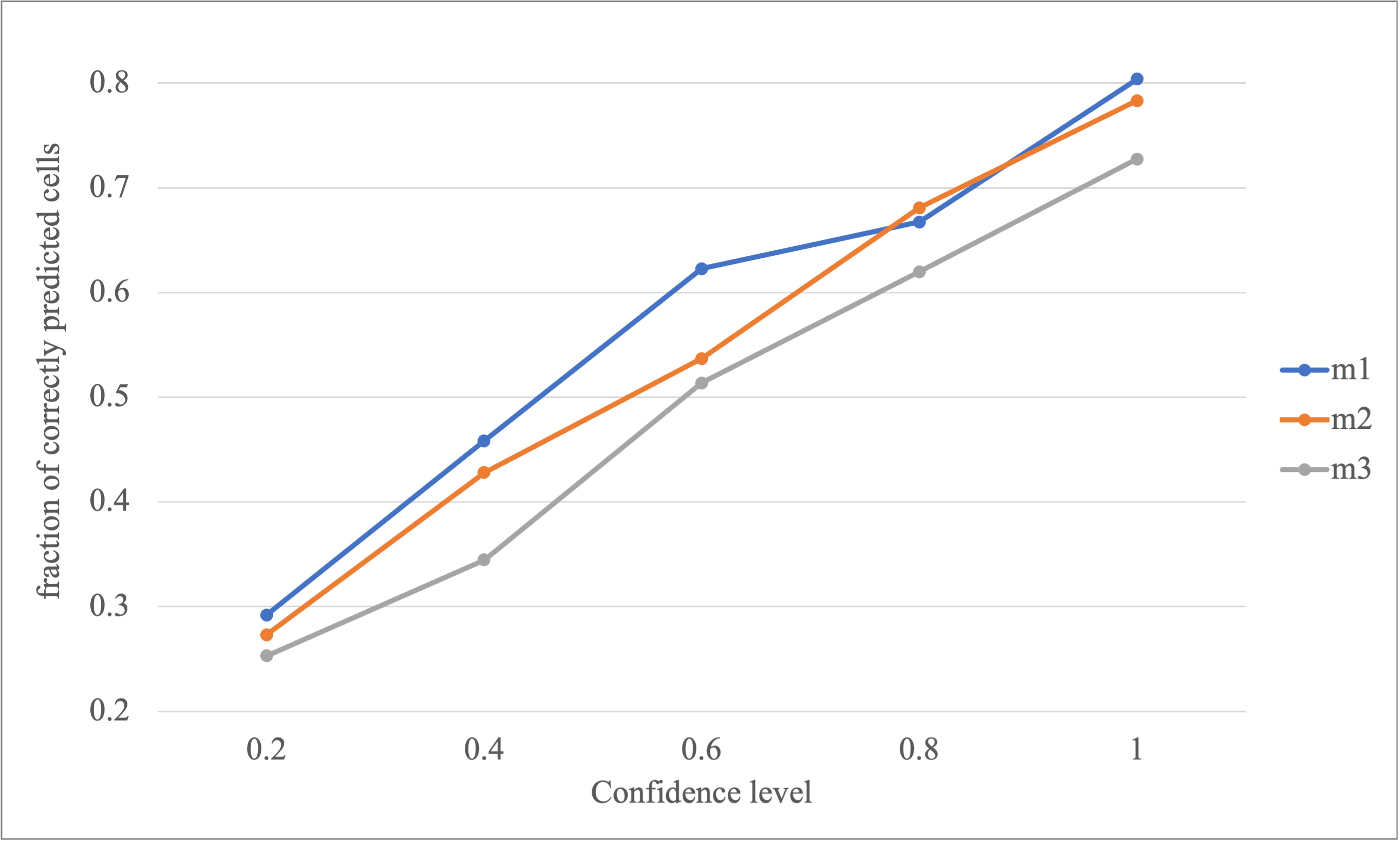}
    \caption{Evaluations of the reliability of the confidence scores as a measure of the uncertainty. m1: no masking applied; m2: pixel values doubled; m3: pixel values tripled. }
    \label{fig:maskeval}
\end{figure}

\paragraph{Experiment 3: Confidence Level as a Measure of Uncertainty Gauged by Pixel Intensity.} The purpose of this experiment is to evaluate proposed models when table image pixel intensity varies. Figure~\ref{fig:maskeval} indicates that the whole curve of the fraction of correctly predicted cells is shifted down as the pixel value increases (so the pixels look fainter). The only exception is when the confidence level is 0.8, but the difference is subtle. Intuitively decreasing pixel intensity (increasing pixel values) should increase the difficulty of accurate detection, leading to higher levels of uncertainty. The results in Figure~\ref{fig:maskeval} indicate that this trend is captured by the confidence levels output by the \texttt{TTA-m} model. 

\paragraph{Experiment~4: Confidence Level as a Measure of Uncertainty Gauged by Cell Complexity.} 
In this experiment, we quantify cell complexity by the adjacency degrees, which is used as a gauge of uncertainty here assuming that cells with more complex structures (and thus adjacency degrees) are likely to be detected with higher uncertainty. Table~\ref{tab:ttam-m-degree} shows that approximately 85\% of the cells in our test data have between 3 to 4 degrees of relationships with neighboring cells. In general, the confidence level decreases as the degree of relationships between cells increases from 1 to 6 with the exception of degree 5. 

\begin{table}
 \caption{Quantifying cell complexity based on the adjacency degree of table cells. The mean confidence level was obtained by taking the average of the confidence scores obtained for all cells for each degree. }
    \centering
    \begin{tabular}{c|c|c|c}
    \toprule
    {\bf Degree} & {\bf \#Cells} & {\bf Cells\%} & {\bf Mean Confidence Level} \\
    \midrule
    {1}  & {27} & {0.5}  & {0.95}  \\
    {2} & {409} & {7.61} & {0.84}    \\
    {3} & {1878} & {34.93} & {0.74}  \\
    {4} & {2937} & {54.62} & {0.71}   \\
    {5} & {115} & {2.14} & {0.77}   \\
    {6} & {8} & {0.15} & {0.65}   \\
    \bottomrule
    \end{tabular}
    \label{tab:ttam-m-degree}
\end{table}

\section{Conclusion and Discussion}
This study explores UQ in TSR problems by modifying the traditional TTA technique and testing it on a customized CascadeTabNet model \cite{prasad2020cascadetabnet}. To evaluate the effectiveness of our UQ method, we used masking and cell complexity quantification techniques. These techniques involve adjusting cell pixel intensity and determining cell complexity based on relationships among cells in table images at different confidence levels. The proposed method demonstrated better Experiments indicating the proposed UQ method provides a more reliable uncertainty estimation.

Compared with the vanilla \texttt{TTA}, \texttt{TTA-m} extends the data augmentation to the training phase, which increases the cost of time to obtain uncertainties of the TSR model. When inferencing the pipeline on a dataset without ground truth labels, one can simply adopt pre-fine-tuned models, so only the test data augmentation is needed. 

Our approach to quantifying uncertainty takes into account both data variation and model variation, unlike the vanilla \texttt{TTA} method that only considers data variation. This is achieved through fine-tuning the target TSR model using our UQ method, which is not limited to any particular TSR model. Additionally, the data augmentation techniques utilized in our study ensure that the TSR model is invariant to different types of tables.

Our study has the following limitations. First, the lack of ground truth limits the capability of assessing real uncertainties. Although we used the pixel intensity and adjacency degree as proxy gauges of detection uncertainties, the real-world data can be hybrid. Such ground truth data could be built by collecting human corrections of automatic annotations by TSR models. Second, the ways we augmented the table images may not be comprehensive. Specifically, the augmentation techniques we explored might not encompass all possibilities or capture the extensive array of variations present in table images. This limitation can be mitigated by building a library of heuristics to modify table images or by building a corpus of artificially synthesized tables. 

\section*{Acknowledgement}
This work has been supported in part by the National Science Foundation (NSF) under Grant Nos. IIS-2245946 and IIS-2236578, 
in part by the Commonwealth Cyber Initiative (CCI),
and in part by the Research Institute of Digital Innovation in Learning (RIDIL).

\bibliographystyle{unsrt}
\bibliography{references}

\end{document}